\documentclass[10pt,twocolumn,letterpaper]{article}

\usepackage{iccv}
\usepackage{times}
\usepackage{epsfig}
\usepackage{graphicx}
\usepackage{amsmath}
\usepackage{amssymb}

\usepackage[breaklinks=true,bookmarks=false]{hyperref}

\iccvfinalcopy 

\def\iccvPaperID{****} 

\ificcvfinal\fi

\begin{document}

\title{Towards Good Practices for Video Object Segmentation}

\author{
Dongdong Yu$^{\dag}$, Kai Su$^{\dag}$, Hengkai Guo, Jian Wang, Kaihui Zhou, Yuanyuan Huang, Minghui Dong, \\ 
Jie Shao and Changhu Wang \\
ByteDance AI Lab, Beijing, China \\
}

\maketitle
\ificcvfinal\thispagestyle{empty}\fi

\let\thefootnote\relax\footnotetext{$^{\dag}$ Equal contribution.}

\begin{abstract}
   Semi-supervised video object segmentation is an interesting yet challenging task in machine learning. In this work, we conduct a series of refinements with the propagation-based video object segmentation method and empirically evaluate their impact on the final model performance through ablation study. By taking all the refinements, we improve the space-time memory networks to achieve a $Overall$ of  $79.1$ on the Youtube-VOS Challenge $2019$.
\end{abstract}

\section{Introduction}
 
 In recent years, video object segmentation  has attracted much attention in the computer vision community~\cite{wug2018fast,perazzi2017learning,khoreva2017lucid,caelles2017one,maninis2018video,shin2017pixel,sun2018mask}.  For a given video, video object segmentation is to classify the foreground and the background pixels in all frames, which is an essential technique for many tasks, such as video analysis, video editing, video summarization and so on. However, video object segmentation is  far from a solved problem, both quality and speed are extremely vital for it.  
 
 The tremendous development of deep convolution neural networks bring huge progress in many areas, including image classification \cite{he2016deep,shen2017multi}, human pose estimation \cite{su2019multi} and video object segmentation~\cite{wug2018fast,perazzi2017learning,khoreva2017lucid,caelles2017one,maninis2018video,shin2017pixel}. These works can be divided into two classes: propagation-based methods~\cite{wug2018fast,perazzi2017learning,khoreva2017lucid} and detection based methods~\cite{caelles2017one,maninis2018video,shin2017pixel}. Propagation based methods, learn a convolution neural network to leverage the temporal coherence of object motion and propagate the mask of the previous frame to current frame. However, there exists some challenging cases, such as occlusions and rapid motion, which cannot be well addressed by the propagation methods. In addition, the propagation error can be accumulated. Detection-based methods, learn the appearance of the target object from a given annotated frame, and perform a pixel-level detection of the target object at each frame. However, they often fail to adapt to appearance changes and have difficulty separating object instances with similar appearances.
 
\begin{figure}[t]
\centering
\includegraphics[width=8cm]{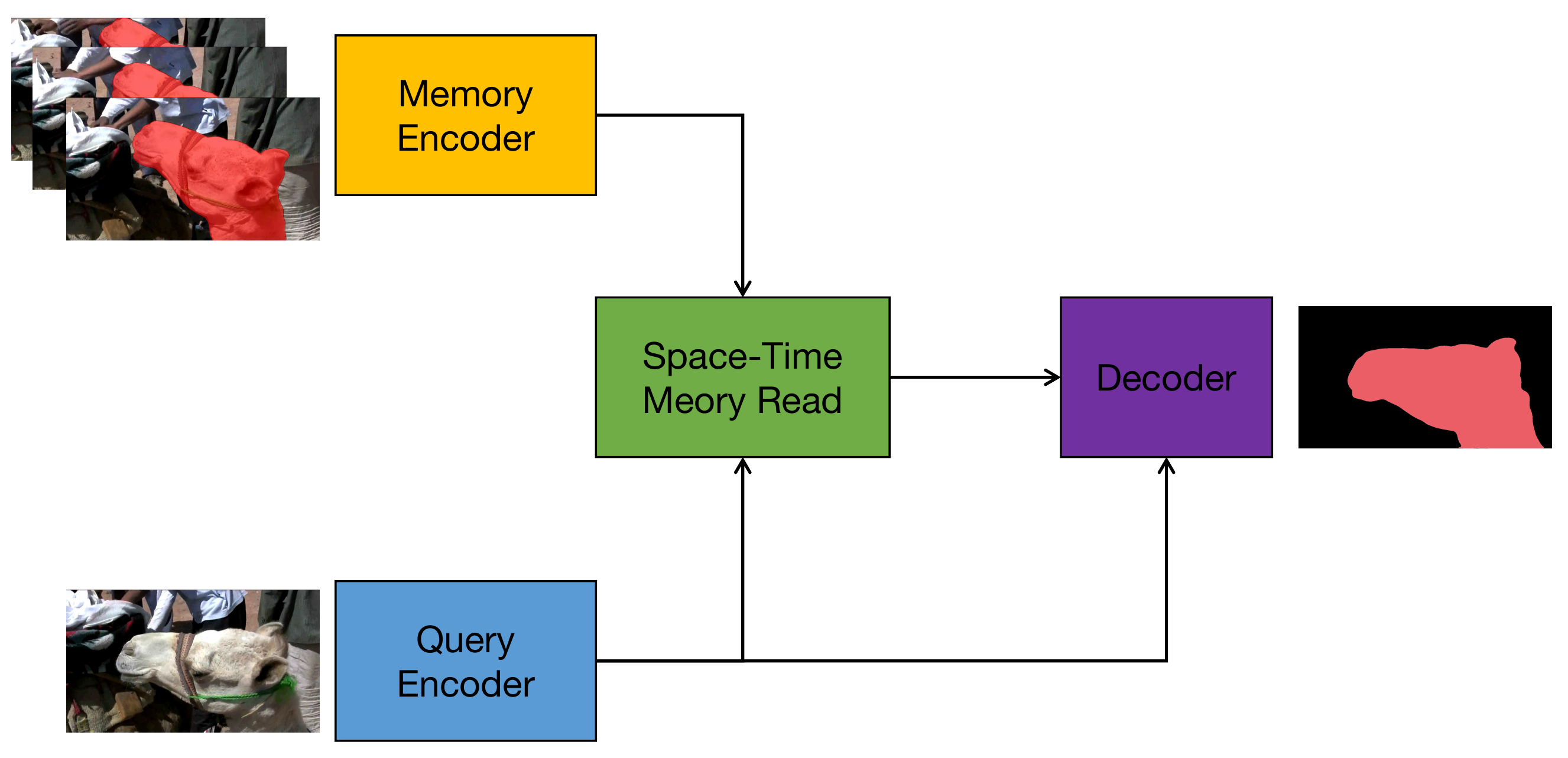}
\caption{Overview of the Space-Time Memory Networks.}
\label{fig:figure1}
\end{figure}
 
 Space-Time Memory Networks~\cite{oh2019video} (STMN) is one of the propagation-based methods, which explores and computes the spatio-temporal attention on every pixel in multiple frames to segment the foreground and the background pixels. By using multi-frame information, it can relieve the bad performance caused by  appearance changes, occlusions, and drifts.  In our paper, we follow STMN and examine a collection of training procedure and model architecture refinements which affect the video object segmentation performance.  First, we explore the segmentation performance of the pre-training stage with different pre-training datasets. Second, we do some ablation study to decide which backbone (including ResNet-$50$, Refine-$50$) should be selected for the encoder. Finally, we validate some testing augmentation tricks, including flip-testing, multi-scale testing and model ensemble, to improve the segmentation performance.

\section{Method}

The chart of Space-Time Memory Networks is shown in Figure \ref{fig:figure1}. During the video processing, the previous frames with object masks are considered as the memory frames and the current frame without the object mask as the query frame. The encoder extracts the appearance information with the memory frames and query frame. The Space-time Memory Read Module will compute the spatio-temporal attention between the query frame and memory frames.  Then, the decoder will output the final segmentation result for the query frame. 

{\bf Pre-training} The STMN is first pre-trained on a simulation dataset generated from static image data, then fine-tuned for real-world videos through the main training. Similar to STMN,  we used image datasets with instance object masks (Pascal VOC \cite{everingham2010pascal, hariharan2011semantic}, COCO\cite{lin2014microsoft}, MSRA10K\cite{shi2016hierarchical}, ECSSD \cite{cheng2014global}, and Youtube-VOS) to simulate training samples. We find that add the Youtube-VOS into the simulation datasets can significantly improve the segmentation performance.

{\bf Backbone:} The STMN use the ResNet-$50$ as the backbone of the encoder and decoder. In our work, we propose a new backbone, named Refine-$50$, which can well handle the scale variant cases.

{\bf Testing Tricks:} In order to improve the segmentation performance, we use the flip-testing and multi-scale testing for a single model. For ensemble experiments, we average the object probability from ResNet-$50$ and Refine-$50$.

\section{Experiments}

In this section, we first briefly introduce the Youtube-VOS \cite{xu2018youtube} dataset and corresponding evaluation metrics, then we evaluate a series of refinements through ablation studies. Finally, we report the final results in the Youtube-VOS Challenge.

\subsection{Datasets and Evaluation Metrics}

Youtube-VOS \cite{xu2018youtube} is the latest large-scale dataset for video object segmentation. The training set consists of $3471$ videos, and we further split the training set into $3321$ offline-training set and $150$ offline-validation set. We adopt the offline-validation set to select the model from different epochs. For evaluation, we measure the region similarity $J$ and contour accuracy $F$. The results of validation set and test set are evaluated through the online CodaLab server.

\subsection{Training Details}

Our model is implemented in Pytorch \cite{paszke2017automatic}. For the training, we $4 V100$ GPUs on a server are used. Adam \cite{kingma2014adam} optimizer is adopted. The learning rate is set to $1e-5$. The input size for the network is made to a fixed $384\times384$. The cross-entropy loss is used. The batch size on each GPU is set to $4$.

\subsection{Testing Details}

Follow \cite{oh2019video}, we simply save a memory frame every $5$ frames. And the input size of the network for inference is set to an integer multiple of $16$. Moreover, we adopt the multi-scale testing to boost the performance.

\subsection{Refinements during Training and Testing Phases}

In this section, we evaluate the effectiveness of a series of refinements during the training and testing phases.

\newcommand{\tabincell}[2]{
	\begin{tabular}{@{}#1@{}}#2\end{tabular}
}
\begin{table}[tb]
	\centering
	\caption{The results of Pre-training, Main-training and Full-training with ResNet-$50$ on YouTube-VOS validation set.}\label{table:pretrain}
	
	\resizebox{0.9\columnwidth}{!}{
	\begin{tabular}{cc}
		\hline
		Training Method & \tabincell{c}{$Overall$} \\
		\hline
		Pre-training only (without Youtube-VOS) & $0.617$ \\
		Pre-training only (with Youtube-VOS) & $0.667$ \\
		Main-training only & $0.681$ \\
		Full-training & $0.766$ \\
		\hline
	\end{tabular}
	}
\end{table}

\subsubsection{Pre-training on images}

We evaluate the performance of different training methods in this experiment. As shown in Table \ref{table:pretrain}, by using Youtube-VOS for pre-training, the performance is improved from 61.7 to 66.7. And, pre-training only achieved perfermance close to main-training only, without adopting any real videos for training. Without the pre-training phase, the performance drops from $0.766$ to $0.681$. Therefore, diverse appearance of different objects during the pre-training stage significantly boost the generalization of our model.

\begin{table}[tb]
	\centering
	\caption{The results of different backbones with pre-training only on YouTube-VOS validation set.}\label{table:backbones}
	
	\resizebox{0.455\columnwidth}{!}{
	\begin{tabular}{cc}
		\hline
		Backbone & \tabincell{c}{$Overall$} \\
		\hline
		ResNet-$50$ & $0.667$ \\
		Refine-$50$ & $0.708$ \\
		\hline
	\end{tabular}
	}
\end{table}

\subsubsection{Different Backbones}

We evaluate the effectiveness of different backbones in this experiment. As shown in Table \ref{table:backbones}, by adopting our stronger refine-$50$ backbone, the results improve from $0.667$ to $0.708$.

\begin{table}[tb]
	\centering
	\caption{The results of flip and multi-scale testing with ResNet-$50$ and full-training on YouTube-VOS validation set.}\label{table:multiscale}
	
	\resizebox{0.6\columnwidth}{!}{
	\begin{tabular}{ccc}
		\hline
		Flip & \tabincell{c}{Multi-Scale} & \tabincell{c}{$Overall$} \\
		\hline
		& & $0.761$ \\
		$\surd$ & & $0.766$ \\
		$\surd$ & $\surd$ & $0.777$ \\
		\hline
	\end{tabular}
	}
\end{table}

\begin{table*}[tb]
	\centering
	\caption{Ranking results on the YouTube-VOS test set.}\label{table:testset}
	
	\resizebox{1.5\columnwidth}{!}{
	\begin{tabular}{cccccc}
		\hline
		Team Name & \tabincell{c}{$Overall$} & \tabincell{c}{$J\_seen$} & \tabincell{c}{$J\_unseen$} & \tabincell{c}{$F\_seen$} & \tabincell{c}{$F\_unseen$} \\
		\hline
		zszhou & $0.818$ & $0.807$ & $0.773$ & $0.847$ & $0.847$ \\
		theodoruszq & $0.817$ & $0.800$ & $0.779$ & $0.833$ & $0.855$ \\
		zxyang1996 & $0.804$ & $0.794$ & $0.759$ & $0.833$ & $0.831$ \\
		swoh & $0.802$ & $0.788$ & $0.759$ & $0.825$ & $0.835$ \\
		Jono & $0.714$ & $0.703$ & $0.680$ & $0.736$ & $0.740$ \\
		andr345 & $0.710$ & $0.699$ & $0.667$ & $0.732$ & $0.740$ \\
		\hline
		Ours (youtube\_test) & $0.791$ & $0.779$ & $0.747$ & $0.815$ & $0.822$ \\
		\hline
	\end{tabular}
	}
\end{table*}

\begin{figure*}[tb]
	\centering
	\includegraphics[scale=1.0]{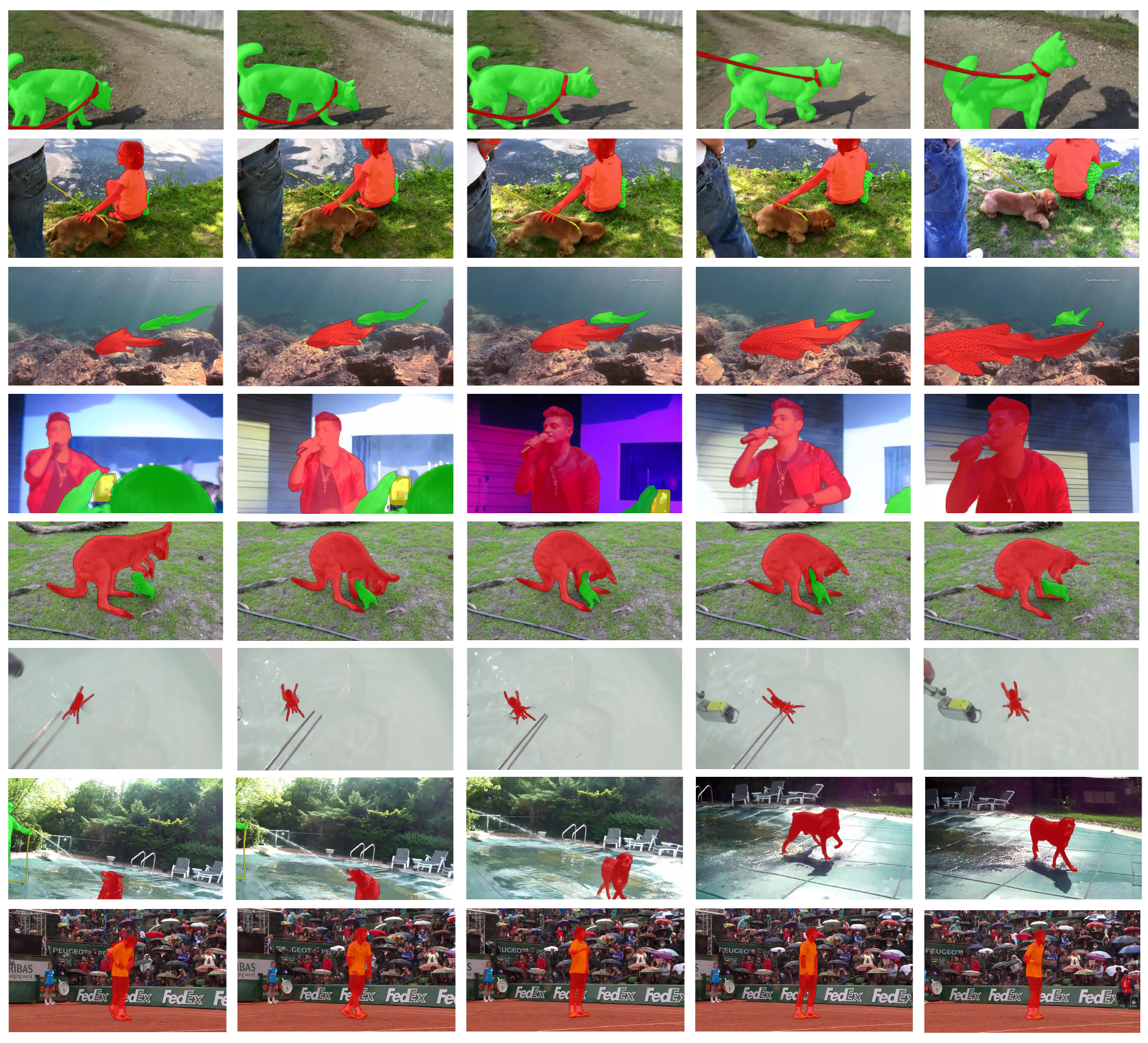}
	\caption{Qualitative results of our model on the YouTube-VOS test set.}
	\label{fig:demo_on_test}
\end{figure*}

\subsubsection{Multi-Scale Testing}

We evaluate the effectiveness of flip and multi-scale testing in this experiment. We adopt the multi-scale with $0.75,1.0$. As shown in Table \ref{table:multiscale}, when adopting the flip testing, the performance improve from $0.761$ to $0.766$. With multi-scale testing involved, we further boost the performance, from $0.766$ to $0.777$.

\subsection{Results on Youtube-VOS Challenge}

Finally, we ensemble the model with ResNet-$50$ and Refine-$50$, and achieved $0.791$ on the Youtube-VOS test set. The qualitative results of the final model are shown in Figure \ref{fig:demo_on_test}.

\section{Discussions}

During our experiments, we find two main problems. Firstly, the results on validation set of the model with different epochs vary seriously. Secondly, the results on validation set and test set for the model with same epoch show a large difference.

\section{Conclusion}

In this work, we conduct a series of refinements with the Space-Time Memory Networks and empirically evaluate their impact on the final model performance through ablation study. Finally, we achieve a $Overeall$ of $79.1$ on the Youtube-VOS Challenge $2019$.

{\small
\bibliographystyle{ieee_fullname}
\bibliography{egbib}
}

\end{document}